\documentclass[10pt,twocolumn,letterpaper]{article}

\usepackage[pagenumbers]{cvpr} 

\usepackage{graphicx}
\usepackage{booktabs}
\usepackage{multirow}
\usepackage{amsmath}

\usepackage{cuted}     
\usepackage{caption}   

\usepackage{amssymb} 
\usepackage{pifont}    
\newcommand{\cmark}{\checkmark}    
\newcommand{\xmark}{\ding{55}}     
\newcommand{\methodname}{UniSH\xspace}
\definecolor{cvprblue}{rgb}{0.21,0.49,0.74}
\usepackage[pagebackref,breaklinks,colorlinks,allcolors=cvprblue]{hyperref}

\def\paperID{} 
\def\confName{CVPR}
\def\confYear{2026}

\title{\methodname: Unifying Scene and Human Reconstruction in a Feed-Forward Pass}

\author{
    Mengfei Li$^{1}$ \quad
    Peng Li$^{1}$ \quad
    Zheng Zhang$^{2}$ \quad
    Jiahao Lu$^{1}$ \quad
    Chengfeng Zhao$^{1}$ \quad
    Wei Xue$^{1}$ \\
    Qifeng Liu$^{1}$ \quad
    Sida Peng$^{3}$ \quad
    Wenxiao Zhang$^{1}$ \quad
    Wenhan Luo$^{1}$ \quad
    Yuan Liu$^{1\dagger}$ \quad
    Yike Guo$^{1\dagger}$ \\[1em]
    $^{1}$The Hong Kong University of Science and Technology \\
    $^{2}$Beijing University of Posts and Telecommunications \quad
    $^{3}$Zhejiang University \\
    {\tt\small mliek@connect.ust.hk, \{yuanly, yikeguo\}@ust.hk}
}

\begin{document}
\maketitle

\renewcommand{\thefootnote}{\fnsymbol{footnote}}
\footnotetext[2]{Corresponding authors.} 
\renewcommand{\thefootnote}{\arabic{footnote}}

\begin{strip}
\vspace{-4em}
    \centering
    \includegraphics[width=1\linewidth]{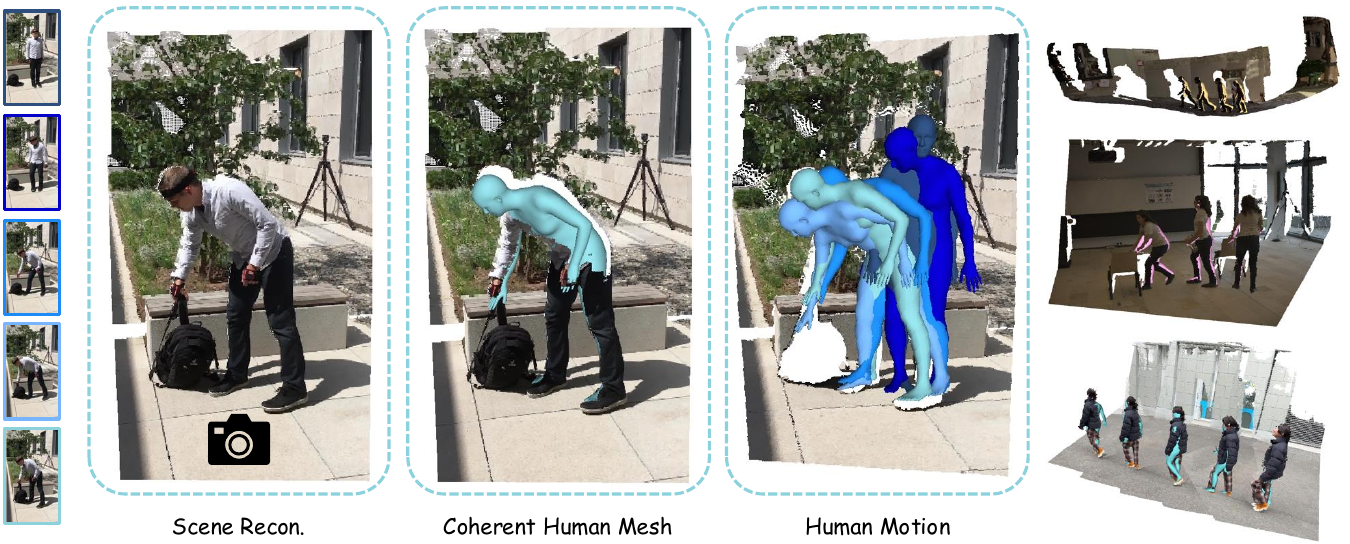}
    \captionof{figure}{Given a monocular video as input, our UniSH is capable of jointly reconstructing scene and human in a single forward pass, enabling effective estimation of scene geometry, camera parameters and SMPL parameters. } 
    \label{fig:teaser}
\end{strip}

\begin{abstract} 

We present \methodname, a unified, feed-forward framework for joint metric-scale 3D scene and human reconstruction. A key challenge in this domain is the scarcity of large-scale, annotated real-world data, forcing a reliance on synthetic datasets. This reliance introduces a significant sim-to-real domain gap, leading to poor generalization, low-fidelity human geometry, and poor alignment on in-the-wild videos. To address this, we propose an innovative training paradigm that effectively leverages unlabeled in-the-wild data. Our framework bridges strong, disparate priors from scene reconstruction and HMR, and is trained with two core components: (1) a robust distillation strategy to refine human surface details by distilling high-frequency details from an expert depth model, and (2) a two-stage supervision scheme, which first learns coarse localization on synthetic data, then fine-tunes on real data by directly optimizing the geometric correspondence between the SMPL mesh and the human point cloud. This approach enables our feed-forward model to jointly recover high-fidelity scene geometry, human point clouds, camera parameters, and coherent, metric-scale SMPL bodies, all in a single forward pass. Extensive experiments demonstrate that our model achieves state-of-the-art performance on human-centric scene reconstruction and delivers highly competitive results on global human motion estimation, comparing favorably against both optimization-based frameworks and HMR-only methods. Project page: \url{https://murphylmf.github.io/UniSH/}


\end{abstract}    
\section{Introduction}


A fundamental goal of computer vision is to enable machines to perceive and understand our complex 3D world. A critical component of this world is human-centric environments, which are defined by the dynamic interplay between static scene geometry and articulated human action. Holistically understanding this interplay, which involves jointly reasoning about the surrounding scene and the people within it, is foundational for the next generation of interactive systems, from augmented and virtual reality to collaborative robotics and embodied AI. These practical demands highlight the need for the joint reconstruction of scene and human.


Most previous works have treated static scene reconstruction~\cite{wang2024dust3r, wang2025continuous, zhuo2025streaming, leroy2024grounding, wang2025vggt, wang2025pi} or Human Mesh Recovery (HMR)~\cite{cai2023smpler, dwivedi2024tokenhmr, goel2023humans, NEURIPS2024_fd23a1f3_neural, Sun_2022_CVPR_BEV, baradel2024multi, kocabas2024pace, li2024coin, Marcard_2018_ECCV_3dpw, Wang_2025_ECCV_tram, Yuan_2022_CVPR_glamr, Ye_2023_CVPR_slamhr} in isolation, a practice that overlooks the essential relationship between human motion and the surrounding 3D environment. More recent research~\cite{muller2025reconstructing, rojas2025hamst3r} has explored mitigating this gap by incorporating multi-view observations and human-centric semantics; however, this line of work primarily focuses on static scenes. Methods such as~\cite{liu2025joint, zhao2024synergistic} are designed for dynamic scenarios but suffer from time-consuming per-scene optimization, limiting their practical applicability. With the advance of the 3R paradigm, exemplified by models like Dust3R~\cite{wang2024dust3r} and Must3R~\cite{cabon2025must3r} which predict a unified 3D signal (camera and scene geometry) in a feed-forward manner, JOSH3R~\cite{liu2025joint} builds on this approach by augmenting the framework with a dedicated human tracking head, enabling both human trajectory prediction and scene reconstruction. Despite this remarkable progress, this approach suffers from limitations, including computationally expensive data annotations, global incoherence, and cumulative errors introduced by its two-frame inference paradigm, which degrade the overall human-scene 4D reconstruction performance.

To perform effective and high quality human-scene 4D reconstruction, we propose UniSH, a feed-forward 4D reconstruction model. UniSH consists of two core branches: a reconstruction branch to recover scene geometry and camera parameters, and a human body branch to estimate human pose and shape. Our framework then fuses features from both branches to regress the absolute metric scale of the scene and the human's global position within it. This entire architecture operates in a single forward pass, yielding a globally-aligned and metric-scale joint reconstruction.


The primary obstacle to training such a model is the profound scarcity of large-scale, annotated real-world data that includes joint 3D scene, human motion, and camera parameters. This forces a reliance on synthetic datasets like BEDLAM~\cite{black2023bedlam}. While synthetic data has proven viable for HMR, it presents critical limitations. First, its limited scene diversity and scale is insufficient for training general-purpose scene reconstruction. Second, training on such data introduces a significant sim-to-real domain gap. Consequently, models trained on this synthetic data generalize poorly when applied to in-the-wild videos, manifesting as degraded scene reconstruction quality and poor SMPL-scene alignment. Furthermore, due to this overall lack of suitable training data, existing reconstruction models have not been adequately optimized for human-centric scenes, resulting in low-fidelity human surface geometry. Therefore, a fundamental challenge is to develop a framework that can effectively leverage abundant, unlabeled real-world data.

To this end, we introduce \textbf{UniSH}, a unified, feed-forward framework for joint metric-scale scene and human reconstruction. Our UniSH leverages the strong priors from state-of-the-art reconstruction ($\pi^3$~\cite{wang2025pi}) and HMR (CameraHMR~\cite{patel2025camerahmr}) models. By building upon these powerful pre-trained foundations, our framework is designed to learn the complex joint reconstruction task even with unlabeled in-the-wild videos and synthetic data. To effectively bridge these distinct priors, we introduce a lightweight module, termed AlignNet, which learns to align the predictions from both models into a single, coherent output.

We also propose a novel training paradigm that effectively utilizes unlabeled, in-the-wild human-centric data. This framework is specifically designed to refine the fidelity of reconstructed human surfaces and to enforce robust SMPL-scene alignment, bridging the gap where synthetic-only supervision fails. To enhance human surface fidelity, we utilize an expert depth model (MoGe-2~\cite{wang2025moge2accuratemonoculargeometry}) to generate pseudo-labels for our unlabeled data and propose a robust supervision strategy to distill high-frequency geometric details. To achieve robust SMPL-scene alignment, we introduce a coarse-to-fine supervision scheme. The model is first coarsely localized using synthetic data. This initialization stabilizes the next alignment stage where visible SMPL is needed. Then, the model is fine-tuned on unlabeled real data by directly optimizing the geometric correspondence between the reconstructed human point cloud and the SMPL mesh, significantly improving real-world generalization. 

Extensive experiments validate our approach. The results demonstrate that UniSH, by integrating powerful reconstruction and HMR priors with our novel training paradigm, successfully achieves robust, joint reconstruction. Our method effectively leverages unlabeled in-the-wild data to bridge the sim-to-real domain gap , delivering high-fidelity human surface details and coherent, metric-scale SMPL bodies accurately aligned to the 3D scene.


Our main contributions are summarized as follows:
\begin{itemize}
    \item A \textbf{unified, feed-forward architecture} that, from a single forward pass, jointly reconstructs high-fidelity scene point clouds, camera parameters, and coherent, metric-scale human meshes.
    


    \item A \textbf{human surface refinement} method that addresses the universal fidelity gap by distilling high-frequency geometric details from a pre-trained expert depth model~\cite{wang2025moge2accuratemonoculargeometry} with unlabeled data.
    
    \item A \textbf{coarse-to-fine supervision scheme} that bridges the sim-to-real domain gap. This strategy first learns coarse localization from synthetic data, then enables fine-grained alignment on unlabeled real-world data by directly optimizing the geometric correspondence between the SMPL mesh and the reconstructed human point cloud.
    
    \item Extensive experiments show our method achieves state-of-the-art human-centric scene reconstruction. On global motion estimation benchmarks, it significantly outperforms prior feed-forward joint reconstruction baselines and remains highly competitive with specialized HMR-only methods and optimization-based joint approaches.
\end{itemize}
\section{Related Work}
\paragraph{3D Scene Reconstruction.}
Traditional Structure-from-Motion (SfM) pipelines~\cite{hartley2003multiple, agarwal2011building, snavely2006photo, snavely2008modeling}, such as the widely adopted COLMAP system~\cite{schonberger2016structure, schonberger2016pixelwise}, follow an incremental paradigm that yields highly accurate sparse geometry but often suffers from high computational cost and failure cases in texture-poor regions or wide-baseline scenarios. In contrast, recent learning-based approaches reconceptualize 3D reconstruction as an end-to-end prediction problem, with DUSt3R~\cite{wang2024dust3r} pioneering dense, pixel-aligned point-map regression and MASt3R~\cite{leroy2024grounding} improving accuracy through an additional dense descriptor head. Building on this foundation, a growing body of work explores online, streaming, and large-scale extensions~\cite{wang2025continuous, chen2025long3r, wang20243d, wu2025point3r, zhuo2025streaming, cabon2025must3r, lu2025align3r, zhang2024monst3r, yang2025fast3r}, broadening applicability in dynamic or memory-constrained settings. More recent feed-forward formulations~\cite{zhang2025flare, wang2025vggt}, like VGGT, leverage transformer architectures to infer point maps for all images in a single pass, eliminating iterative optimization. $\pi^3$~\cite{wang2025pi} further improves the model's precision and generalization by introducing the permutation-equivariant setting. 




\paragraph{3D Human Reconstruction.}
Parallel to scene reconstruction, 3D human reconstruction has evolved from early camera-centric mesh estimation toward recovering full-body pose and shape within a global, world-grounded frame. Modern Human Mesh Recovery (HMR) is built on parametric body models such as SMPL~\cite{loper2023smpl} and SMPL-X~\cite{pavlakos2019expressive}, which encode human geometry using low-dimensional pose and shape parameters; methods like SMPLify~\cite{bogo2016keep} estimate these parameters by optimizing reprojection consistency with 2D joints. Enabled by large-scale datasets~\cite{black2023bedlam, dwivedi2024tokenhmr, patel2021agora, feng2024chatpose, joo2021exemplar}, learning-based approaches~\cite{dwivedi2024tokenhmr, kanazawa2018end, li2022cliff, omran2018neural, zhang2021pymaf} directly regress parametric model parameters in a single forward pass, while recent Transformer-based architectures~\cite{vaswani2017attention} further improve robustness and accuracy by exploiting powerful vision-transformer backbones~\cite{dosovitskiy2020image} for single-person mesh recovery~\cite{goel2023humans, patel2025camerahmr, wang2023refit, xu2023smpler, cai2023smpler, yin2025smplest}. Beyond parametric forms, another line of research reconstructs high-fidelity non-parametric human surfaces~\cite{xiu2022icon, xiu2023econ, zhang2023globalcorrelated, ho2024sith, Zhang_2024_CVPR, xue2024human3diffusion, li2024pshuman, chen2025synchuman}, typically aiming to recover complete meshes including non-visible regions within a camera-centric coordinate system. Although our work also refines human geometry, its focus fundamentally differs: we refine only the visible human surface and prioritize accurate absolute localization and alignment within the global 3D scene.

\paragraph{Joint Scene-Human Reconstruction.}

Early work on joint reconstruction~\cite{pavlakos2022one}, such as HSfM~\cite{muller2025reconstructing}, employed optimization-based pipelines for static, multi-view environments. HAMSt3R~\cite{rojas2025hamst3r} advanced this direction but remained limited to static scenes and required a separate SMPL fitting stage. The critical limitation of these methods is their failure to generalize to dynamic, monocular input. To address this, subsequent optimization-based methods~\cite{zhao2024synergistic, xue2024hsr, liu2025joint} tackled dynamic scenes. SyncHMR~\cite{zhao2024synergistic} leverages SMPL priors for metric-scale SLAM, while JOSH~\cite{liu2025joint} uses scene geometry to physically ground human motion. A principal limitation of these approaches, however, is their reliance on time-consuming per-scene optimization. To overcome this, JOSH3R~\cite{liu2025joint} introduced a feed-forward approach by augmenting a scene backbone with a human tracking head. While a significant advance, its two-frame paradigm can introduce global incoherence and cumulative errors. Concurrent work, Human3R~\cite{chen2025human3r}, also moves toward a unified system by building upon an online 4D model, CUT3R~\cite{wang2025continuous}, for direct motion regression.

\section{Method}

\begin{figure*}[ht]
    \centering
    \includegraphics[width=1\linewidth]{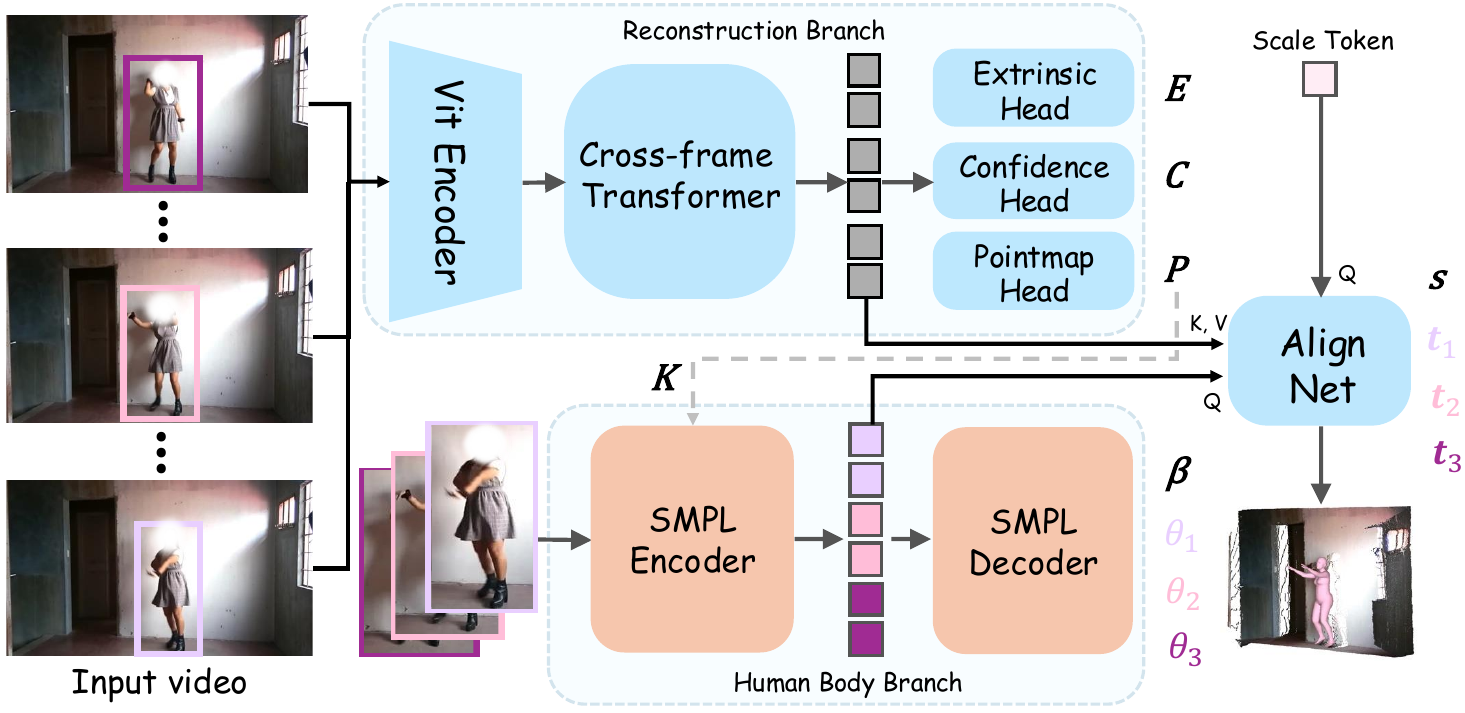}
    \caption{\textbf{The network architecture of UniSH.} UniSH takes a monocular video as input. The video frames are processed by the \textbf{Reconstruction Branch} to predict per-frame camera extrinsics $E$, confidence maps $C$, and pointmaps $P$. Camera intrinsics $K$ are derived from the pointmaps. Human crops from the video are fed into the \textbf{Human Body Branch} along with $K$ to estimate global SMPL shape parameters $\beta$ and per-frame pose parameters $\theta_i$. Features from both branches are processed by \textbf{AlignNet} to predict the global scene scale $s$ and per-frame SMPL translations $t_i$ for coherent scene and human alignment. The subscript $i$ denotes the frame index.}
    \label{fig:pipeline}
\end{figure*}

\methodname takes a sequence of $N$ images $\mathcal{I}=\{I_i\}_{i=1}^N$ as input. In a single forward pass, it predicts the per-frame metric-scale point map $\mathcal{P}=\{P_i\}_{i=1}^N$ of both human and scene, along with corresponding camera poses $\mathcal{E}=\{E_i=[R_i|T_i]\}_{i=1}^N$ and intrinsics $\mathcal{K}=\{K_i\}_{i=1}^N$. For humans in the scene, our model regresses camera-frame SMPL pose $\theta_{i}$, and position $t_{i}$, as well as a shared body shape $\beta$ across frames.


\subsection{Model Architecture}
As detailed in Fig.~\ref{fig:pipeline}, \methodname includes a scene reconstruction branch, a human body regression branch, and an \textbf{AlignNet} for human-scene alignment.

\vspace{-1em}
\paragraph{Scene Reconstruction.}


The scene reconstruction branch builds upon $\pi^3$~\cite{wang2025pi}, a feed-forward and permutation-equivariant 3D reconstruction model. Given an image sequence $\mathcal{I}=\{I_i\}_{i=1}^N$, an image encoder and a cross-frame encoder are first employed to extract a unified geometric feature $\mathcal{F}_{\text{geo}}$, which is then processed by three decoders to predict per-frame camera poses~$\mathbf{\mathcal{E}}$, point maps~$\mathbf{\mathcal{P}}$, and associated confidence maps~$\mathbf{\mathcal{C}}$, respectively. The intrinsics $\mathcal{K}$ can be subsequently derived from the predicted point maps. We utilize a similar architecture to leverage the strong structural priors of $\pi^3$.


\paragraph{Human Mesh Reconstruction.} 
We adopt CameraHMR~\cite{patel2025camerahmr} as the human body branch, owing to its strong generalization capabilities for single-frame pose and shape estimation. For each frame, a human bounding box $b_i$ is first extracted using an off-the-shelf detector~\cite{Shen_2025_SA_gvhmr}. The bounding box, together with the focal length (derived from $K_i$) and the input image, are provided to the human body branch to extract per-frame body features $\mathcal{F}_{\text{hmr}}$ and output the per-frame pose parameters $\theta_i$ and the shared shape $\beta$.

\paragraph{Scale Prediction and SMPL Placement.}
A naive approach to transitioning the reconstruction branch from an unknown-scale to a metric-scale predictor is to supervise the reconstruction branch with synthetic data to directly predict metric-scale geometry and SMPL placement. However, this strategy is ineffective (Fig.~\ref{fig:ablation}a), as it severely degrades the powerful structural priors of the pre-trained model and exhibits poor generalization to in-the-wild data due to the sim-to-real domain gap. To address these issues, we introduce AlignNet, a lightweight two-layer transformer decoder designed to predict a global scene scale $s$ and the per-frame SMPL translations $\mathcal{T}$, which jointly align the human mesh to the metric-scale scene. We formulate AlignNet as a transformer architecture and take the geometric features $\mathcal{F}_{\text{geo}}$ from the scene reconstruction branch as key-value pairs, while the query is formed by concatenating a learnable scale token $T_s$ with the sequence of per-frame HMR tokens $\mathcal{F}_{\text{hmr}}$. Formally,
$$
(s, \mathcal{T}) = \text{AlignNet}(\mathcal{F}_{\text{geo}}, [T_s | \mathcal{F}_{\text{hmr}}])
$$
where $\mathcal{T} = \{t_i\}_{i=1}^N$ represents the set of per-frame translations, and $[\cdot | \cdot]$ denotes concatenation.



\subsection{Human Surface Refinement}
A primary challenge in general-purpose scene reconstruction is the resulting low-fidelity human surface geometry. A straightforward approach to enhance this is to fine-tune the reconstruction branch on synthetic data (e.g., BEDLAM~\cite{black2023bedlam}) using its original training objective. However, we find this strategy is detrimental. Due to the significant sim-to-real domain gap, this fine-tuning degrades performance on real-world inputs, damaging the model's powerful pre-trained structural priors.

To address this, we adopt an inherently scalable approach, curating a new large-scale video dataset of unlabeled, in-the-wild human-centric motion (see Supp. Mat.). We further employ MoGe-2~\cite{wang2025moge2accuratemonoculargeometry}, a state-of-the-art monocular depth estimator, to produce pseudo ground-truth depth maps. However, direct supervision using these predictions is non-trivial. The per-frame depth estimates lack temporal consistency, exhibiting inconsistent global scales, and are affected by noise, particularly around background regions and human boundaries. 

\paragraph{Confidence-Aware Local Human Loss.}
Inspired by the local geometry enhancement strategy in MoGe-2~\cite{wang2025moge2accuratemonoculargeometry}, we propose a novel confidence-aware, human-centric local loss to robustly distill high-frequency surface details from the pseudo-depth labels. Specifically, for an input frame, we sample $K$ anchor pixels located on the foreground human mask $M_i$ obtained via SAM2~\cite{ravi2024sam2}. $X_k$ is the corresponding 3D coordinate projected from the pseudo-depth label. For each anchor point $X_k$, we define a local patch of neighboring points $\mathcal{N}_k$ by a threshold $\tau$:
$$
\mathcal{N}_k = \{X_k^j \mid \|X_k^j - X_k\|_2 \le \tau\}
$$
Let $\{ \hat{D}_k^j \}$ and $\{ C_k^j \}$ be our model's predicted depths and confidences for the $j^{th}$points $X_k^j \in \mathcal{N}_k$, and let $\{ D_k^j \}$ be their corresponding pseudo-depth labels. To resolve the local scale and shift ambiguity inherent in monocular depth estimation, we employ an ROE Solver~\cite{wang2025moge2accuratemonoculargeometry} for each patch. This solver estimates an optimal scale $s_k$ and shift $t_k$ that best align our model's predictions $\{ \hat{D}_k^j \}$ to the pseudo-depth labels $\{ D_k^j \}$ within the patch $\mathcal{N}_k$. For a given frame $I_i$, the local human loss $\mathcal{L}_{h,i}$ is then formulated as a confidence-weighted $L_1$ distance:
$$
\mathcal{L}_{h,i} = \frac{1}{K} \sum_{k=1}^K \left( \frac{1}{|\mathcal{N}_k|} \sum_{j=1}^{|\mathcal{N}_k|} C_k^j \cdot |(s_k \cdot \hat{D}_k^j + t_k) - D_k^j| \right)
$$

To prevent catastrophic forgetting, we also apply a regularization loss $\mathcal{L}_{\text{preg}}$ that penalizes the deviation from the original pre-trained model's output point map $P_i^{\text{orig}}$. The final objective for Stage 1 can be formulated as:
\begin{equation}
\label{eq:stage1}
\mathcal{L}_{\text{stage1}} = \frac{1}{N} \sum_{i=1}^N \left( \lambda_{h} \mathcal{L}_{h,i} + \lambda_{\text{preg}} \| P_i - P_i^{\text{orig}} \|_1 \right)
\end{equation}

We train the point map decoder while keeping the rest parameters frozen.

\subsection{Human-Scene Alignment}
A straightforward approach to this alignment task is to train the model to predict scale and SMPL translation using only synthetic data. However, due to the aforementioned sim-to-real domain gap, models trained in this way generalize poorly to in-the-wild scenarios~(Fig.~\ref{fig:ablation}b). We therefore design a coarse-to-fine supervision scheme, where a coarse alignment stage first learns a robust initial SMPL placement. This initialization is essential, as it enables the subsequent fine-grained alignment stage which relies on SMPL projection.

\paragraph{Coarse-grained Alignment.}
In this stage, we use synthetic data BEDLAM \cite{black2023bedlam} to learn an initial localization. We supervise the model's predicted translation $t_i$ against the ground-truth $t_i^{\text{gt}}$ using a globally optimal scale $s_{\text{opt}}$ (estimated with an ROE Solver \cite{wang2025moge2accuratemonoculargeometry}). The per-frame coarse alignment loss, $\mathcal{L}_{\text{smpl}, i}$, is defined as:
\begin{equation}
\label{eq:smpl_coarse_i}
\begin{aligned}
\mathcal{L}_{\text{smpl}, i} = & \lambda_{\text{v}} \|V_{\text{smpl}, i} - V_{\text{smpl}, i}^{\text{gt}}\|_1 + \lambda_{\text{j3d}} \|J_{\text{3d}, i} - J_{\text{3d}, i}^{\text{gt}}\|_1 \\
& + \lambda_{\text{j2d}} \|J_{\text{2d}, i} - J_{\text{2d}, i}^{\text{gt}}\|_1 + \lambda_{\text{pose}} \|\theta_i - \theta_i^{\text{gt}}\|_2^2 \\
& + \lambda_{\text{shape}} \|\beta - \beta^{\text{gt}}\|_2^2 + \lambda_{\text{trans}} \|s_{\text{opt}} \cdot t_i - t_i^{\text{gt}}\|_2^2
\end{aligned}
\end{equation}
This loss combines standard supervision signals for HMR, including losses on vertices ($V_{\text{smpl}, i}$), 3D keypoints ($J_{\text{3d}, i}$), 2D projected keypoints ($J_{\text{2d}, i}$), and the SMPL parameters ($\theta_i, \beta, t_i$). The 2D keypoints $J_{\text{2d}, i}$ are obtained by projecting $J_{\text{3d}, i}$ using the predicted intrinsics $K_i$: $J_{\text{2d}, i} = \Pi(K_i, J_{\text{3d}, i})$.

In this stage, we freeze the weights of the fine-tuned reconstruction branch and train both the human body branch and AlignNet. The total loss $\mathcal{L}_{\text{stage2}}$ combines the per-frame loss $\mathcal{L}_{\text{smpl}, i}$ (Eq. \ref{eq:smpl_coarse_i}) with direct supervision on the predicted global scale $s$ against $s_{\text{opt}}$:
\begin{equation}
\label{eq:stage2}
\mathcal{L}_{\text{stage2}} = \frac{\lambda_{\text{smpl}}}{N}\sum_{i=1}^N \mathcal{L}_{\text{smpl}, i} + \lambda_{\text{scale}} \| s - s_{\text{opt}} \|_1
\end{equation}

\paragraph{Fine-grained Alignment.}
In fine-grained alignment stage, we address the generalization gap by fine-tuning the model on unlabeled in-the-wild data. This is achieved by introducing new objectives that directly minimize the geometric error between the predicted SMPL mesh and the reconstructed human point cloud for each frame. This fine-grained alignment relies on a geometric alignment loss and a depth-ordering regularization term.

We first obtain the predicted SMPL mesh then use a differentiable renderer to acquire a visibility mask $M_{\text{vis}, i}$ for this mesh. The source point cloud $V_{\text{src}, i}$ is then defined as the set of visible vertices $V_{\text{smpl}, i}[M_{\text{vis}, i}]$. The target point cloud $V_{\text{tgt}, i}$ is derived from the reconstructed human geometry. It is obtained by scaling the per-frame point map $P_i$ by the predicted scale $s$ and filtering it with human mask $M_{i}$ from SAM2~\cite{ravi2024sam2}: $V_{\text{tgt}, i} = (s \cdot P_i)[M_{i}]$.

The primary alignment loss is the one-way Chamfer distance from the visible SMPL vertices to the target point cloud:
\begin{equation}
\label{eq:align}
\mathcal{L}_{\text{align}, i} = \sum_{v_{\text{src}} \in V_{\text{src}, i}} \min_{v_{\text{tgt}} \in V_{\text{tgt}, i}} \| v_{\text{src}} - v_{\text{tgt}} \|_2^2
\end{equation}

Additionally, we introduce a depth-ordering regularization term. We define $\bar{d}_{\text{src}, i}$ and $\bar{d}_{\text{tgt}, i}$ as the mean depths (Z-values) of the source and target points, respectively. This loss enforces the physical prior that the reconstructed human point cloud (which should be closer to the camera) should not be occluded by the underlying SMPL mesh:
\begin{equation}
\label{eq:depth_reg}
\mathcal{L}_{\text{dreg}, i} = \text{ReLU}(\bar{d}_{\text{tgt}, i} - \bar{d}_{\text{src}, i})
\end{equation}

Finally, in this third stage, we fine-tune AlignNet only with our unlabeled real-world data. The total loss $\mathcal{L}_{\text{stage3}}$ combines the fine-grained alignment loss $\mathcal{L}_{\text{align}, i}$ (Eq.~\ref{eq:align}), the depth-ordering regularization $\mathcal{L}_{\text{dreg}, i}$ (Eq.~\ref{eq:depth_reg}), and a 2D keypoint reprojection loss $\mathcal{L}_{\text{j2d}, i}$. This 2D loss is supervised using pseudo-annotations from a pre-trained CameraHMR~\cite{patel2025camerahmr}.
\begin{equation}
\label{eq:stage3}
\begin{aligned}
\mathcal{L}_{\text{stage3}} = \frac{1}{N} \sum_{i=1}^N \Big( & \lambda_{\text{align}} \mathcal{L}_{\text{align}, i} + \lambda_{\text{depth}} \mathcal{L}_{\text{dreg}, i}
 + \lambda_{\text{j2d}} \mathcal{L}_{\text{j2d}, i} \Big)
\end{aligned}
\end{equation}

\begin{figure*}
    \centering
    \includegraphics[width=1\linewidth]{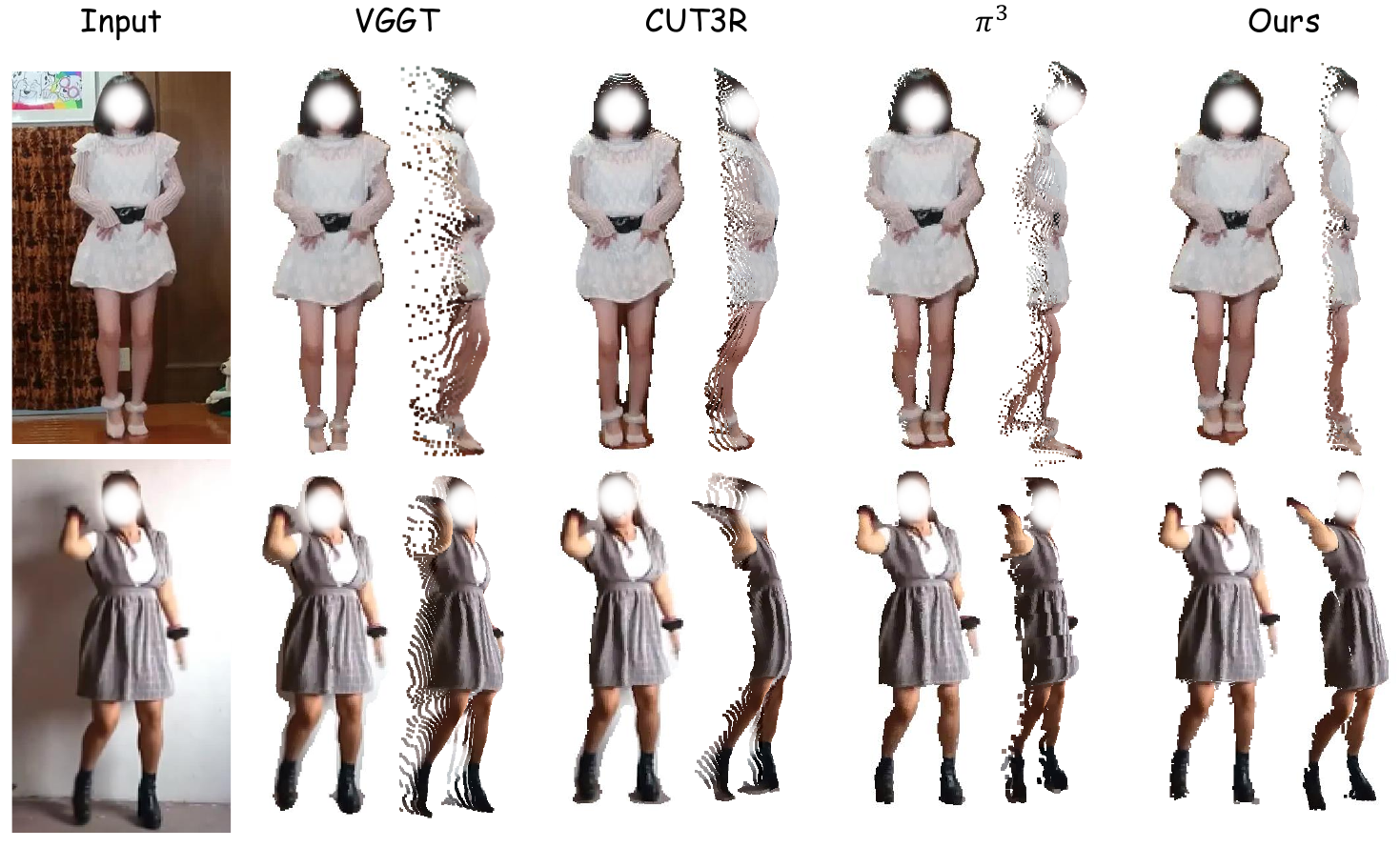}
    \caption{\textbf{Qualitative comparisons of human point cloud.} With in-the-wild input, we compare the reconstructed human point cloud with strong reconstruction model baselines. Benefit from our surface refinement strategy, our UniSH generates consistently better human surface point cloud than all baseline methods.}
    \label{fig:pointcloud_result}
\end{figure*}

\begin{figure*}
    \centering
    \includegraphics[width=1\linewidth]{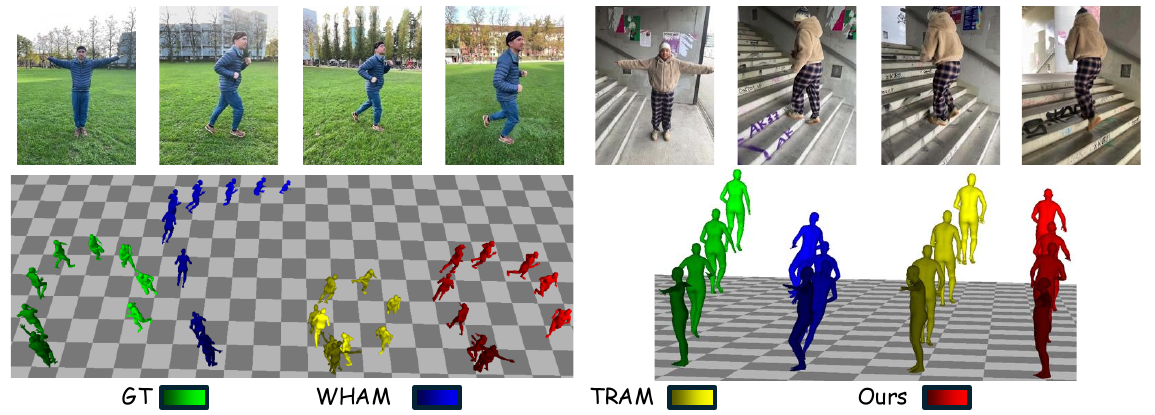}
    \caption{\textbf{Qualitative results of global human motion estimation.} We compare our method with well known HMR methods WHAM~\cite{Shin_2024_CVPR_wham} and TRAM~\cite{Wang_2025_ECCV_tram} on EMDB-2. Our method shows competitive results to these methods that specially designed and optimized for global human motion estimation task.}
    \label{fig:global_vis}
\end{figure*}
\section{Experiments}

In this section, we present the empirical validation for our proposed methodology. Our evaluation protocol is designed to assess two primary capabilities. We first benchmark our approach against leading methods in general 3D reconstruction, specifically on a human-centric depth estimation task. We then analyze the model's accuracy in estimating global human motion within a world-coordinate system. Finally, a detailed ablation study is provided to elucidate the specific contributions of our framework's key components.

\subsection{Human-centric Scene Reconstruction}
We first evaluate our framework's reconstruction performance on human-centric video depth estimation, a task demanding high per-frame accuracy and temporal consistency. Following prior protocols~\cite{wang2025pi}, we evaluate on the Bonn dataset~\cite{palazzolo2019refusion} using two standard metrics: Absolute Relative Error (Abs Rel $\downarrow$) and thresholded accuracy ($\delta < 1.25 \uparrow$).

As shown in Table~\ref{tab:bonn_scale}, our method quantitatively outperforms all leading baselines. \methodname achieves an Abs Rel of \textbf{0.035} and a $\delta < 1.25$ accuracy of \textbf{0.980}, significantly surpassing strong methods like $\pi^{3}$~\cite{wang2025pi} (0.049/0.975) and VGGT~\cite{wang2025vggt} (0.057/0.966). This quantitative leap is supported by qualitative results from in-the-wild videos (Fig.~\ref{fig:pointcloud_result}). These comparisons demonstrate that our method reconstructs human surfaces with richer details and more plausible geometry than prior reconstruction-focused approaches, validating the synergistic benefits of our human-centric training.

\begin{table}[t]
\centering
\begin{tabular}{lcc}
\toprule
Method & Abs Rel $\downarrow$ & $\delta < 1.25 \uparrow$ \\
\midrule
DUSt3R~\cite{wang20243d}     & 0.151  & 0.839  \\
MASt3R~\cite{leroy2024grounding}     & 0.188  & 0.765  \\
MonST3R~\cite{zhang2024monst3r}     & 0.072  & 0.957  \\
Fast3R~\cite{yang2025fast3r}     & 0.194  & 0.772  \\
MVDUSt3R~\cite{tang2025mv}   & 0.426  & 0.357  \\
CUT3R~\cite{wang2025continuous}       & 0.078  & 0.937  \\
Aether~\cite{team2025aether}     & 0.273  & 0.594  \\
FLARE~\cite{zhang2025flare}       & 0.152  & 0.790  \\
VGGT~\cite{wang2025vggt}       & 0.057  & 0.966  \\
$\pi^{3}$~\cite{wang2025pi} & 0.049  & 0.975  \\
\textbf{UniSH(Ours)}                      & \textbf{0.035}   &  \textbf{0.980}  \\
\bottomrule
\end{tabular}
\caption{\textbf{Quantitative results of human-centric video depth estimation on the Bonn~\cite{palazzolo2019refusion} dataset.} Our approach, \methodname, significantly outperforms all prior reconstruction-focused baselines.}
\label{tab:bonn_scale}
\end{table}

\begin{table*}[t!]
\centering
\small 
\begin{tabular}{l|c|c|ccc|ccc}
\toprule
\multirow{2}{*}{\textbf{Method}} & \multirow{2}{*}{\begin{tabular}[c]{@{}c@{}}\textbf{Opt.}\\\textbf{Free}\end{tabular}} & \multirow{2}{*}{\textbf{Scene}} & \multicolumn{3}{c|}{\textbf{EMDB-2~\cite{kaufmann2023emdb}}} & \multicolumn{3}{c}{\textbf{RICH~\cite{Hassan_2019_ICCV_rich}}} \\
\cmidrule(lr){4-6} \cmidrule(lr){7-9}
& & & WA-MPJPE $\downarrow$ & W-MPJPE $\downarrow$ & RTE(\%) $\downarrow$ & WA-MPJPE $\downarrow$ & W-MPJPE $\downarrow$ & RTE(\%) $\downarrow$ \\
\midrule
SLAHMR~\cite{Ye_2023_CVPR_slamhr} & \xmark & \xmark & 326.9 & 776.1 & 10.2 & 132.2 & 237.1 & 6.4 \\
TRAM~\cite{Wang_2025_ECCV_tram} & \xmark & \xmark & 76.4 & 222.4 & 1.4 & 127.8 & 238.0 & 6.0 \\
JOSH~\cite{liu2025joint} & \xmark & \cmark & \textbf{68.9} & \textbf{174.7} & \textbf{1.3} & \textbf{89.0} & \textbf{132.5} & \textbf{3.0} \\
\midrule
TRACE~\cite{Sun_2023_CVPR_trace} & \cmark & \xmark & 429.0 & 1702.3 & 17.7 & 238.1 & 925.4 & 101.4 \\
WHAM~\cite{Shin_2024_CVPR_wham} & \cmark & \xmark & 135.6 & 334.8 & 6.0 & 108.4 & 190.1 & 4.5 \\
GVHMR~\cite{Shen_2025_SA_gvhmr} & \cmark & \xmark & \textbf{111.0} & \textbf{276.5} & \textbf{2.0} & \textbf{78.8} & \textbf{126.3} & \textbf{2.4} \\
\midrule
JOSH3R~\cite{liu2025joint} & \cmark & \cmark & 220.0 & 661.7 & 13.1 & - & - & - \\
\textbf{UniSH(Ours)} & \cmark & \cmark & \textbf{118.5} & \textbf{270.1} & \textbf{5.8} & 118.1&183.2 & 4.8\\
\bottomrule
\end{tabular}
\caption{\textbf{Evaluation of global human motion estimation on EMDB-2 and RICH datasets.} We categorize methods by their properties: \textbf{Opt. Free}($\cmark$) indicates whether the method is feed-forward; \textbf{Scene}($\cmark$) indicates the method jointly reconstructs 3D scene geometry. }
\label{tab:global_hmr}
\end{table*}

\subsection{Global Human Motion Estimation}

We next evaluate the model's accuracy in estimating global human motion on the EMDB-2~\cite{kaufmann2023emdb} and RICH~\cite{Hassan_2019_ICCV_rich} datasets. Following prior protocols~\cite{Shin_2024_CVPR_wham, Wang_2025_ECCV_tram}, we divide each sequence into 100-frame segments and evaluate 3D joint errors (mm) using three key metrics. World-aligned MPJPE (WA-MPJPE) performs a single Procrustes alignment over the entire segment to assess motion shape consistency. World MPJPE (W-MPJPE) aligns only the first two frames, evaluating drift. We also report Root Translation Error (RTE, \%), the error after rigid alignment, to quantify long-term trajectory accuracy.

We present quantitative results in Table~\ref{tab:global_hmr}. While HMR specialized methods achieve superior numerical accuracy, \methodname offers a distinct set of advantages. Methods like JOSH~\cite{liu2025joint} are optimization-based and computationally expensive. In contrast, specialized feed-forward methods like GVHMR~\cite{Shen_2025_SA_gvhmr} are trained on extensive ground-truth data and, crucially, do not reconstruct the 3D scene. Our approach is deliberately designed as a feed-forward, joint-reconstruction framework that relies primarily on weaker supervision from unlabeled in-the-wild data. This trade-off is an expected outcome. Given this, our model achieves highly competitive results, significantly outperforming JOSH3R~\cite{liu2025joint}, the other feed-forward, joint-reconstruction baseline. Crucially, \methodname is the only method presented that is simultaneously feed-forward, and jointly reconstructs the 3D scene. This joint approach provides a key qualitative advantage, shown in Fig.~\ref{fig:global_vis}; by leveraging strong structural priors from $\pi^3$ for robust camera pose recovery, our method achieves competitive results against WHAM and TRAM, particularly in ill-posed scenarios.

\subsection{Ablation Study}

\begin{figure*}
    \centering
    \includegraphics[width=1\linewidth]{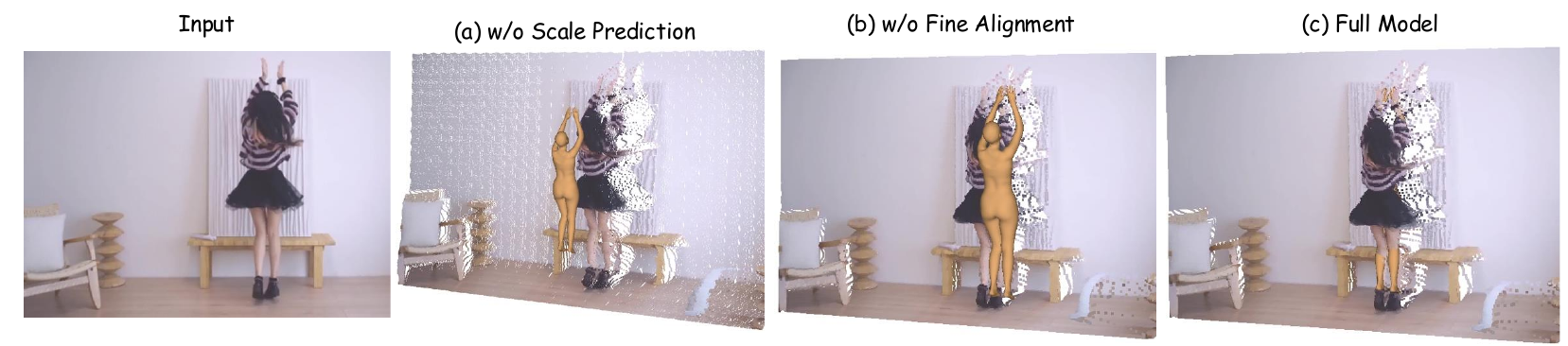}
    \caption{\textbf{Ablation study of our key design.} (a) A variant where the scene branch is directly supervised for metric scale, and the Align Net only predicts SMPL translation. (b) Our model trained with only the coarse (synthetic) alignment stage, omitting the fine-grained alignment. (c) Our full model, which incorporates both coarse (synthetic) and fine-grained (real-world) alignment stages.}
    \label{fig:ablation}
\end{figure*}

\paragraph{Surface Refinement.}
We validate our human surface refinement strategy in Table~\ref{tab:ablation_bonn}. Compared to the $\pi^3$ baseline ('No Refinement'), naively fine-tuning on synthetic BEDLAM data significantly degrades performance, confirming the sim-to-real domain gap. While a mixed 'BEDLAM + Real' training still underperforms the baseline, our 'Real (Ours)' approach achieves the best performance by a substantial margin. This validates that our distillation strategy, using only unlabeled in-the-wild data, is crucial for bridging the domain gap and successfully refining high-fidelity human geometry.

\paragraph{Pipeline Design.}

We qualitatively ablate our pipeline design in Figure~\ref{fig:ablation} by comparing three variants.
\textbf{(a)} Directly supervising the scene branch with BEDLAM's GT metric scale (bypassing scale prediction is detrimental, as it severely damages the powerful $\pi^3$ prior, degrades reconstruction quality, and fails to recover metric scale on real-world data.
\textbf{(b)} Training with only the coarse (synthetic) alignment stage leads to poor generalization; due to the sim-to-real domain gap, the SMPL mesh and scene remain poorly aligned on in-the-wild videos.
\textbf{(c)} Our full model resolves this issue, as the fine-grained alignment stage leverages unlabeled real-world data to optimize geometric correspondence, achieving robust human-scene alignment and validating our full two-stage, coarse-to-fine scheme.

\begin{table}
\centering
\caption{\textbf{Ablation study of our human surface refinement on the Bonn~\cite{palazzolo2019refusion} dataset.} 'BEDLAM' denotes fine-tuning using only ground-truth (GT) supervision from the synthetic BEDLAM dataset.  'Real' means using only real-world data and our proposed surface refinement loss.}
\label{tab:ablation_bonn}
\begin{tabular}{lcc}
\toprule
Training Data & Abs Rel $\downarrow$ & $\delta < 1.25 \uparrow$ \\
\midrule
No Refinement & 0.049 & 0.975 \\
BEDLAM & 0.062 & 0.960 \\
BEDLAM + Real & 0.051 & 0.968 \\
\textbf{Real(Ours)} & \textbf{0.035} & \textbf{0.980} \\
\bottomrule
\end{tabular}
\end{table}
\section{Conclusion}

We have proposed \methodname, a unified feed-forward framework for joint, metric-scale scene and human reconstruction. Our core contribution is a novel training paradigm that addresses the critical sim-to-real domain gap by leveraging unlabeled, in-the-wild data. By distilling geometric cues from expert models and fine-tuning with a coarse-to-fine alignment strategy, \methodname successfully integrates strong, disparate priors from scene reconstruction and Human Mesh Recovery. As a result, our method produces coherent scene geometry, accurate camera parameters, and globally-aligned SMPL bodies in a single pass.

While \methodname achieves robust human-scene alignment, the resulting non-parametric human geometry can still exhibit artifacts, such as floaters. This limitation suggests a clear direction for future work: investigating how the aligned SMPL bodies, provided by our model, can be used as a strong geometric prior to regularize and refine the final human surface.
{
    \small
    \bibliographystyle{ieeenat_fullname}
    \bibliography{main}
}

\clearpage
\setcounter{page}{1}
\maketitlesupplementary

\subsection{Dataset Curation}
\label{sec:dataset_curation}

\paragraph{Scalability and Data Requirements.}
A core advantage of UniSH is the ability to learn from unlabeled in-the-wild videos with minimal geometric assumptions. For the dataset curated in this work, we specifically target dance videos to capture high-quality, complex human motion where the subject remains the primary focus. However, this domain choice serves primarily as a proof-of-concept. Unlike prior work that necessitates laboratory settings or motion capture, our framework imposes no such restrictions. We only require monocular videos with a visible human subject. Consequently, our method can be effortlessly scaled up to significantly more diverse, general-purpose video sources available on the public internet.

\paragraph{Automated Filtering Pipeline.}
We curated a large-scale dataset from public platforms, employing a rigorous, multi-stage automated pipeline to ensure geometric and temporal consistency. First, to mitigate the impact of scene transitions and montages common in raw internet videos, we enforce strict temporal continuity. We employ the standard PySceneDetect library to identify shot boundaries based on content changes. Sequences exhibiting such discontinuities are automatically discarded, ensuring that the input to our temporal attention modules represents a continuous, unedited motion sequence.

Following temporal filtering, we employ the object detector adopted by~\cite{Shen_2025_SA_gvhmr} to isolate single-subject sequences. We retain only those clips where a unique person class is consistently detected, thereby eliminating crowd ambiguity. To guarantee sufficient resolution for surface refinement, we enforce a spatial prominence constraint, requiring the subject's average bounding box height to exceed 40\% of the image height.

Finally, visibility is enforced by discarding sequences with bounding box truncation at image borders. Crucially, to ensure an unobstructed view, we filter out any sequence where the subject overlaps with other detected bounding boxes, effectively removing environmental occlusions. The resulting dataset comprises 1,354 unique sequences, totaling approximately 1.2 million frames.

\paragraph{Ethical Compliance.}
Our data collection protocol strictly aligns with the conference ethical guidelines. Acknowledging the impracticability of obtaining individual consent given the dataset scale, we restricted our acquisition exclusively to content publicly broadcast by original creators. We operate strictly within the scope of the fair use doctrine for academic research. To safeguard subject privacy, we enforce a policy against the retention of any personally identifiable metadata.

Furthermore, we uphold the right to be forgotten through a passive distribution mechanism. Rather than distributing raw video data, we release only Video IDs and corresponding timestamps. This ensures that any content removed by the creator from the hosting platform automatically becomes inaccessible within our dataset. This mechanism effectively preserves the creator's ultimate control over their content dissemination.

\paragraph{Bias and Limitations.}
We acknowledge potential biases in the data source. Online dance communities may skew towards specific demographics or body types. Users should be aware of these distribution shifts during deployment. However, our core contribution is the methodology for leveraging abundant in-the-wild videos. Since our framework requires no manual labels, this specific bias is not a limitation of the method itself. It can be readily mitigated by simply scaling the data collection to include more diverse sources.

\section{Model Architecture Details}

This section details the unified architecture of the UniSH framework. Our model integrates three specialized components. These modules collaborate to achieve joint metric-scale scene and human reconstruction in a single forward pass.

\subsection{Scene Reconstruction Branch}

This branch adapts the permutation-equivariant architecture of $\pi^3$~\cite{wang2025pi}. The goal is permutation-equivariant geometry estimation. A ViT-Large encoder~\cite{dosovitskiy2020image} is adopted as the backbone. The core feature aggregation employs a Cross-Frame Transformer. This Transformer uses 36 layers. The hidden dimension is $D=1024$. It is configured with 16 attention heads. Feature tokens alternate between spatial and global self-attention. The architecture omits frame index positional embeddings. Three parallel heads process the geometric features $\mathcal{F}_{geo}$. These heads predict the extrinsic camera poses $E$, the per-frame point map $P$, and the confidence map $C$.

\subsection{Human Body Branch}

We adopt the CameraHMR~\cite{patel2025camerahmr} framework for human pose and shape estimation, utilizing a ViTPose-Base backbone to extract image features. The architecture employs a Transformer Decoder that processes the input human crop. To condition the reconstruction on scene-specific geometry, we inject bounding box and focal length information as done in~\cite{patel2025camerahmr}. Crucially, the focal length $f$ is not estimated latently but is derived directly from the intrinsics predicted by our Scene Reconstruction Branch, explicitly coupling the two branches.

The SMPL Decoder outputs the per-frame pose parameters $\theta_i$ and the body shape parameters $\beta \in \mathbb{R}^{10}$. To ensure temporal consistency across the sequence, we compute the final body shape by averaging the per-frame $\beta$ predictions over the entire video. In addition to the parametric outputs, the decoder extracts high-level human feature tokens $\mathcal{F}_{hmr}$, which serve as the query input for the subsequent AlignNet module.

\subsection{AlignNet}

The AlignNet is our novel lightweight fusion network. Its primary function is aligning the human and scene predictions into a single metric-scale coordinate system.

The AlignNet is implemented as a two-layer transformer decoder with a default hidden dimension of $D_{hidden}=512$. It operates with dedicated adapter layers. These layers map the input geometric features $\mathcal{F}_{geo}$ and human features $\mathcal{F}_{hmr}$ to the uniform hidden dimension $D_{hidden}$.

The query sequence $Q$ is explicitly constructed. $Q$ is formed by concatenating a learned scale token with the sequence of adapted Human Body features $\mathcal{F}_{hmr}$. The key $K$ and value $V$ sequences are derived from the consolidated Scene Geometry features $\mathcal{F}_{geo}$.

The prediction layers consist of a stack of two Cross-View Transformer Decoder layers. These layers utilize rotary position embeddings to encode the temporal sequence information.

The decoder output is used by two prediction heads. The dedicated Scale Head processes the Scale Token output. The predicted global scale $s$ is obtained by applying the softplus activation function to the logits, ensuring positivity. The Translation Head predicts the camera translation $t_i$. The final translation vector $t_i$ is constructed by scaling the raw $(x, y)$ components by the final depth component $z_{final}$, then concatenating with $z_{final}$.

\section{Implementation Details}

This section outlines the specific numerical configurations and optimization routines used to train UniSH. We detail the general setup and the critical weighting parameters for each training stage.

\subsection{General Optimization and Training Setup}

The entire framework is trained using the AdamW optimizer with $\beta_1=0.9$ and $\beta_2=0.95$. We employ a cosine decay learning rate policy. The initial learning rate is $5 \times 10^{-5}$ for all three training stages. Training runs for a total of 100 epochs per stage. We utilize a linear warm-up phase for the first 3 epochs, followed by the cosine decay schedule, decaying the learning rate to zero. We use a distributed training setup. The per-GPU batch size is configured to 2. We use gradient accumulation over 4 steps. This yields an effective total batch size of 64 with 8 GPUs. The weight decay parameter for regularization is fixed at $0.05$. In each iteration, we sample a video clip spanning 5 seconds. This clip is temporally sampled to 30 frames at 6 FPS for training. The input image resolution is $518 \times 294$ pixels. Gradient Clipping is enabled with a maximum gradient norm of 1.0.

\subsection{Weights of Losses}

\paragraph{Stage 1: Human Surface Refinement} This stage is designed to distill high-frequency geometric details from an expert depth model using unlabeled, in-the-wild videos. The training objective for this stage is given by:
$$
\mathcal{L}_{stage1}=\frac{1}{N}\sum_{i=1}^{N}(\lambda_{h}\mathcal{L}_{h,i}+\lambda_{preg}||P_{i}-P_{i}^{orig}||_{1})
$$
The primary weight for the confidence-aware local human loss is $\lambda_{h} = 1$. The regularization weight penalizing deviation from the original pre-trained point map is $\lambda_{preg} = 0.1$. The local patch radius $\tau$ for computing $\mathcal{L}_{h,i}$ is set to 0.2.

\paragraph{Stage 2: Coarse-grained Alignment} This stage establishes initial metric-scale consistency using synthetic data. The coarse alignment loss, $\mathcal{L}_{stage2}$, combines HMR supervision with a global scale loss:
$$
\mathcal{L}_{stage2}=\frac{\lambda_{smpl}}{N}\sum_{i=1}^{N}\mathcal{L}_{smpl,i}+\lambda_{scale}||s-s_{opt}||_{1}
$$

The weights are $\lambda_{smpl} = 1$ and $\lambda_{scale} = 0.1$. The component loss weights are set as $\lambda_{v} = 0.1$, $\lambda_{j3d} = 0.1$, $\lambda_{j2d} = 10$, $\lambda_{pose} = 0.1$, $\lambda_{shape} = 0.1$, and $\lambda_{trans} = 1$.

\paragraph{Stage 3: Fine-grained Alignment} This final stage addresses the generalization gap by fine-tuning on unlabeled real-world data. The objective directly minimizes the geometric error between the reconstructed human point cloud and the predicted SMPL mesh. The total loss is:
$$
\mathcal{L}_{stage3}=\frac{1}{N}\sum_{i=1}^{N}(\lambda_{align}\mathcal{L}_{align,i}+\lambda_{depth}\mathcal{L}_{dreg,i}+\lambda_{j2d}\mathcal{L}_{j2d,i})
$$
We set the alignment weight $\lambda_{align} = 1$, the depth ordering regularization weight $\lambda_{depth} = 1$, and the $2D$ reprojection loss weight $\lambda_{j2d} = 10$.

\section{Impact of Human Surface Refinement}

\begin{figure*}[t]
    \centering
    \includegraphics[width=\linewidth]{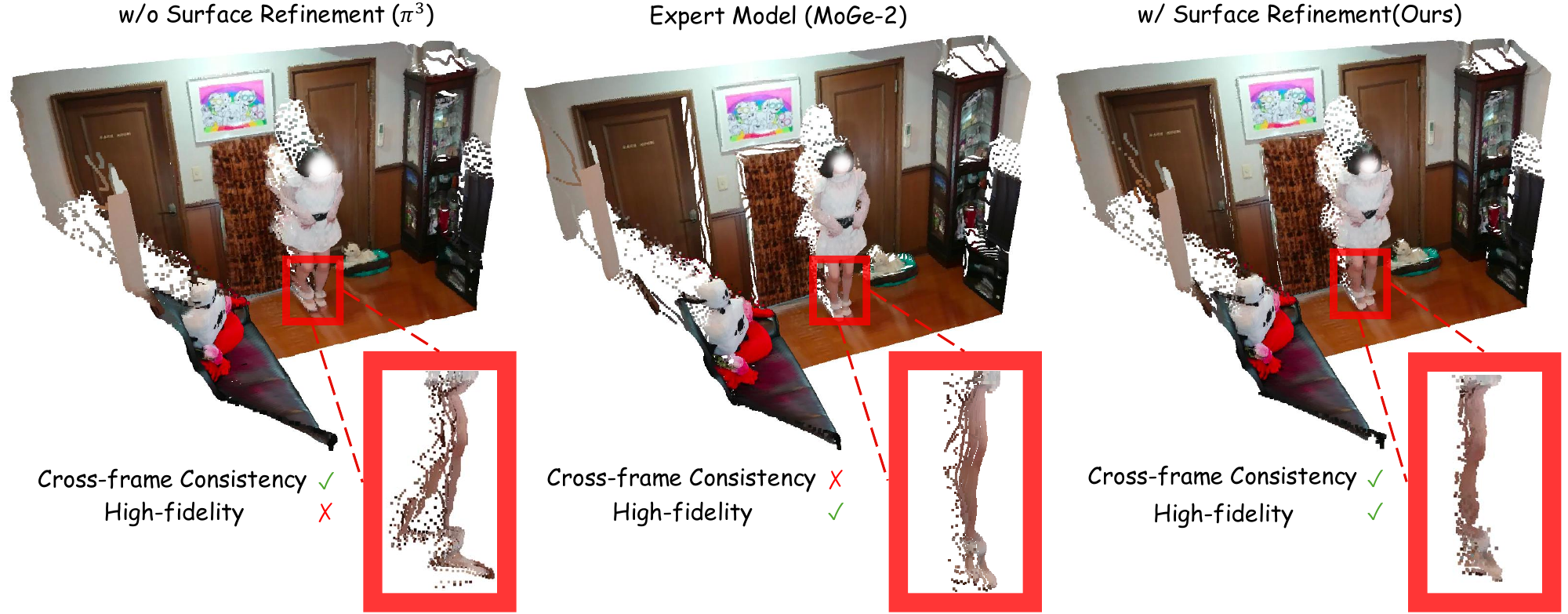}
    \caption{\textbf{Qualitative Impact of Human Surface Refinement.} Comparison illustrating the effectiveness of our specialized surface refinement strategy. The $\pi^3$ baseline (Left) provides good cross-frame consistency but the reconstructed human geometry is coarse and does not conform well to the body shape. The Expert Monocular Model (MoGe-2) (Center) achieves high fidelity but lacks multi-view consistency. Our full method (Right) successfully distills the high-frequency surface details into the multi-view reconstruction framework, achieving both high fidelity and strong cross-frame consistency.}
    \label{fig:surface_refinement_impact}
\end{figure*}

\begin{figure*}[ht]
    \centering
    \includegraphics[width=1.0\linewidth]{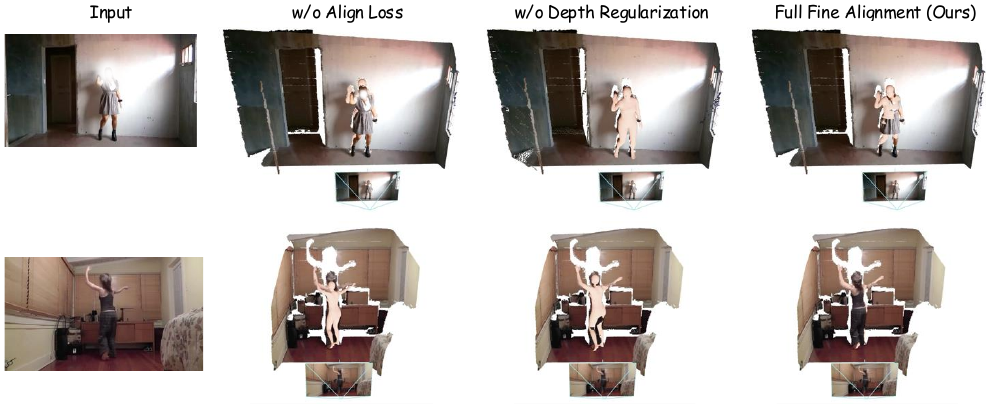} 
    \caption{\textbf{Visual ablation of the Fine-grained Alignment stage.} 
    We validate the necessity of our unsupervised geometric losses on in-the-wild data. 
    \textbf{w/o Align Loss:} Without explicit geometric alignment ($\mathcal{L}_{align}$), the model fails to predict the correct scene scale and SMPL placement due to the domain gap.
    \textbf{w/o Depth Regularization:} Removing the depth constraint ($\mathcal{L}_{dreg}$) results in physical inconsistencies, where the SMPL mesh is not correctly positioned to align with the visible human point cloud.
    \textbf{Full Fine Alignment (Ours):} Our complete method achieves accurate global alignment and correct depth ordering.
    \textbf{Note:} We omit the visualization for ``w/o Coarse Alignment'' as removing the initial coarse stage leads to training non-convergence.}
    \label{fig:fine_align_ablation}
\end{figure*}

Our surface refinement strategy is designed to overcome the inherent trade-off between geometric fidelity and multi-view consistency. The baseline $\pi^3$ model~\cite{wang2025pi}, shown on the left, provides strong cross-frame consistency for the scene and coarse human structure. However, the resulting human point cloud often exhibits poor geometric fidelity. It fails to accurately conform to the detailed shape of the human body.

Conversely, a specialized monocular depth estimator, such as the expert MoGe-2 model~\cite{wang2025moge2accuratemonoculargeometry} (center), can predict exceptionally high-fidelity human surface details on a per-frame basis. This approach, however, fundamentally suffers from poor temporal consistency and global scene alignment.

Our method successfully integrates these dual requirements (Fig.~\ref{fig:surface_refinement_impact}). By employing confidence-aware distillation from the MoGe-2 expert, UniSH effectively injects high-frequency geometric information into the robust $\pi^3$ framework. The result is a system that maintains the global coherence and cross-frame consistency inherited from the $\pi^3$ backbone, while simultaneously achieving significantly higher fidelity and detail in the reconstructed human surface. This qualitative synthesis validates the necessity and efficacy of our specialized two-stage human surface refinement approach.


\section{Impact of Fine-grained Alignment}
\label{sec:fine_grain_alignment}
Our framework employs a coarse-to-fine strategy to align the reconstructed human mesh with the scene. We first emphasize the necessity of this curriculum. The initial coarse alignment stage is a strict prerequisite for convergence. Attempting to optimize the geometric fine-tuning objectives directly from random initialization results in training collapse.

Once initialized by the coarse stage, the fine-grained alignment handles the critical sim-to-real adaptation. We visually ablate the components of this stage in Figure \ref{fig:fine_align_ablation}. The second column shows the performance without the geometric alignment loss ($\mathcal{L}_{align}$). It fails to generalize to in-the-wild inputs due to the domain gap. This results in incorrect global scale predictions and erroneous SMPL translations that drift from the visual evidence.

The third column demonstrates the model trained without the depth-ordering regularization ($\mathcal{L}_{dreg}$). The alignment loss successfully pulls the SMPL mesh near the human point cloud. However, the lack of physical depth constraints causes the SMPL body to be placed incorrectly. It often floats in front of the reconstructed surface. This violates the physical principle that the camera-visible surface should occlude the internal body volume. Our full model successfully integrates both objectives. It achieves accurate metric scale and translation while maintaining correct depth ordering. The SMPL mesh is placed coherently to align with the reconstructed human surface.

\section{More Visualization Results}

\begin{figure*}[t]
    \centering
    \includegraphics[width=\linewidth]{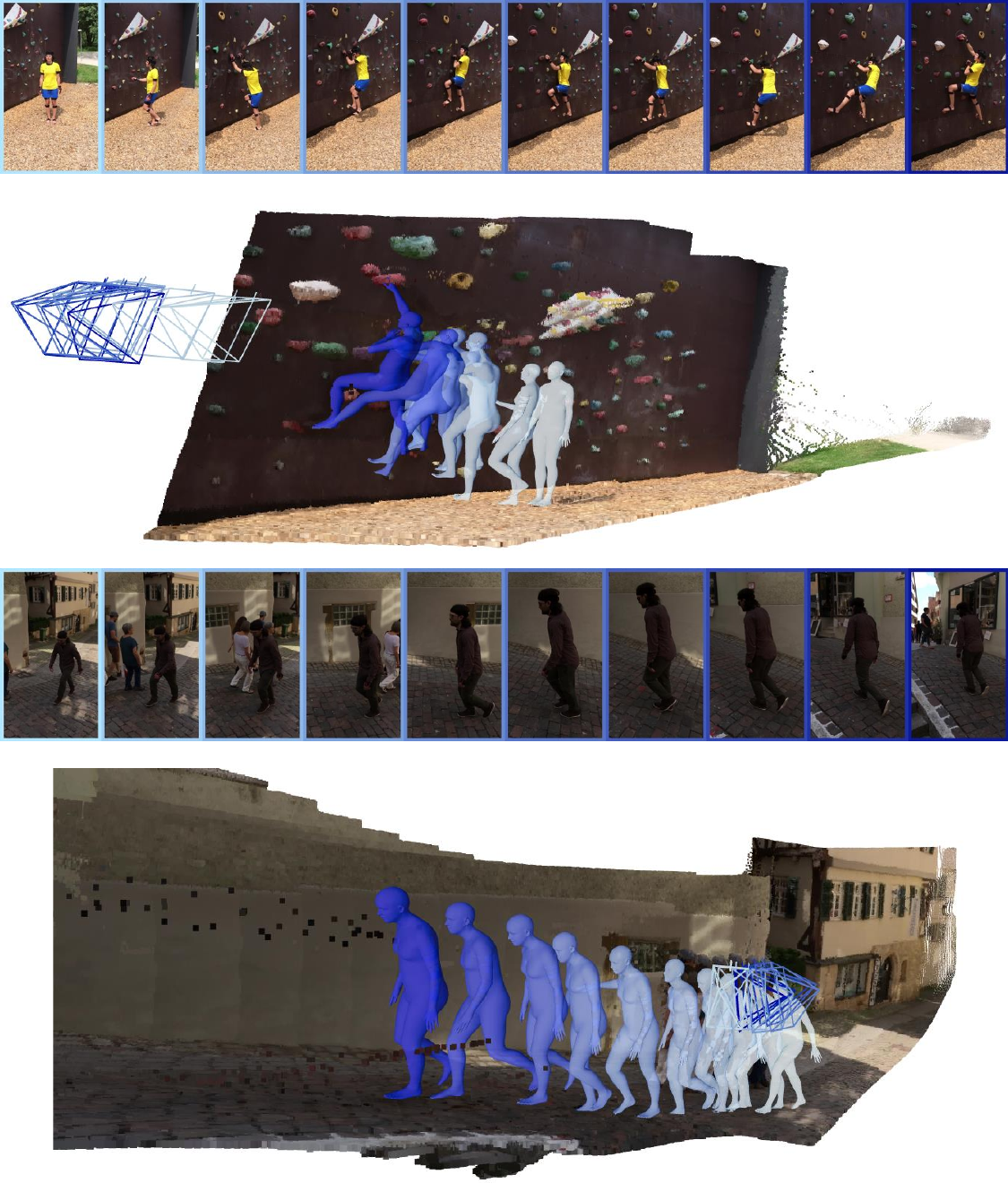}
    \caption{\textbf{Qualitative Visualization of Joint Scene and Human Reconstruction.} The examples demonstrate the robustness and metric consistency of our framework. The color gradient (light blue to dark blue) consistently encodes the temporal sequence across both the reconstructed camera poses and the SMPL meshes. The upper example illustrates robustness in reconstructing highly articulated poses (rock climbing) and accurately aligning the SMPL mesh with the scene geometry. The lower example demonstrates coherent, long-term tracking of human motion in a complex urban environment, verifying the metric stability and generalization of our joint reconstruction framework.}
    \label{fig:more_visualizations}
\end{figure*}

This section presents additional qualitative results demonstrating the robustness of UniSH across diverse, challenging scenes and complex human motions. In Fig.~\ref{fig:more_visualizations}, the temporal sequence is explicitly encoded by the color gradient, ranging from light blue (earlier frames) to dark blue (later frames), which is consistently applied to the reconstructed cameras and the SMPL meshes.

The upper sequence showcases reconstruction of an extreme, highly articulated human pose (rock climbing). Our framework accurately aligns the SMPL body model to the reconstructed scene geometry (the climbing wall), confirming stability even under non-standard articulation. The lower sequence illustrates coherent, long-term tracking of human motion within a large, in-the-wild urban environment. The color-coded trajectory clearly shows the consistent movement of both the predicted SMPL meshes and their corresponding camera poses over time. The framework successfully reconstructs the complex irregular wall structure and maintains stable metric-scale alignment of the human trajectory over multiple frames. These visualizations confirm the generalization capability and metric stability of the joint scene and human reconstruction provided by UniSH.

\section{4D Visualizations}
We provide a supplementary video file to further illustrate our method. This video focuses on two primary aspects of our performance. First, we demonstrate the temporal consistency of the UniSH framework. We show the dynamic evolution of the reconstructed scene and human mesh on continuous video sequences. This highlights the stability of our method in maintaining geometric coherence over time. 

Second, we include dynamic visualizations of our ablation study. We compare our full model against the ablated variants discussed in Section \ref{sec:fine_grain_alignment}. This video comparison provides a clearer perspective on how our alignment strategies function in dynamic scenarios compared to static image figures. Please refer to the attached file \texttt{demo\_video.mp4}.

\end{document}